\documentclass[11pt]{article}
\usepackage{coling2020}
\usepackage{times}
\usepackage{url}
\usepackage{latexsym}

\usepackage[utf8]{inputenc} 
\usepackage[colorinlistoftodos]{todonotes}
\usepackage{multirow}
\usepackage[toc,page]{appendix}

\title{Transfer Learning of Lexical Semantic Families \\for Argumentative Discourse Units Identification}

\author{João A. Rodrigues \and Ruben Branco \and Ant{\'o}nio Branco \\
  University of Lisbon \\
  NLX -- Natural Language and Speech Group, Department of Informatics \\
  Faculdade de Ciências, Campo Grande, 1749-016 Lisboa, Portugal \\
  \texttt{\{jarodrigues, rmbranco, ambranco\}@fc.ul.pt}
}

\date{}

\begin{document}
\maketitle
\begin{abstract}

Argument mining tasks require an informed range of low to high complexity linguistic phenomena and commonsense knowledge.
Previous work has shown that pre-trained language models are highly effective at encoding syntactic and semantic linguistic phenomena when applied with transfer learning techniques and built on different pre-training objectives.
It remains an issue of how much the existing pre-trained language models encompass the complexity of argument mining tasks.
We rely on experimentation to shed light on how language models obtained from different lexical semantic families leverage the performance of the identification of argumentative discourse units task.
Experimental results show that transfer learning techniques are beneficial to the task and that current methods may be insufficient to leverage commonsense knowledge from different lexical semantic families.


\end{abstract}

\section{Introduction}
\label{intro}

The language technology landscape has been changed by pre-trained language models (LMs) obtained from semantic spaces of words \cite{mikolov2013distributed}, sentences \cite{kiros2015skip} and now shifting towards contextual words \cite{devlin2019bert}.
Although these pre-trained LMs have brought huge improvements, the encoded knowledge and transferability~\cite{caruana1997multitask} are still to be fully understood. 

Argument mining \cite{lawrence2020argument}, the automatic identification of argumentative structures within natural language expressions, has improved and gained new insights resorting to pre-trained LMs.
Recent work has incorporated pre-trained LMs using transfer learning techniques to predict argument structure~\cite{pathak2016two,li2017word}, convincingness~\cite{simpson2018finding}, or identify argumentative component relations~\cite{cocarascu2017identifying,rocha2018cross}.
Transfer learning techniques have also been successful across-domains for the classification of argumentative discourse units~\cite{Khatib2016CrossDomainMO,ajjour2017unit} and with multi-task transfer~\cite{Eger2017NeuralEL,Potash2017HeresMP,Stab2018CrosstopicAM,choi2018gist,schulz2018multi,jo2019cascade}.

We extend prior work by addressing argument mining linguistic complexity with different lexical semantic families.
We resort to three distinct data sets and perform a range of analysis regarding the identification of argumentative discourse units (ADUs)\cite{peldszus2013argument} task.
We aim to answer the question: What is the impact of models pre-trained with different lexical semantic families on ADUs identification?
We address this question by evaluating several classifiers pre-trained with different semantic spaces and provide insights over the linguistic knowledge that encompasses the task.

\section{Methodology}

\textbf{Task and Data sets} - We address the identification of ADUs with the following data sets: UKP Sentential Argument Mining Corpus (UKPS) \cite{Stab2018CrosstopicAM}, Webis-Debate-16 (WEBIS) \cite{Khatib2016CrossDomainMO} and AraucariaDB (ARAUC) \cite{Reed2008LanguageRF}.
These mainstream data sets are annotated with different and somewhat competing criteria.
The data sets were adapted for the identification of ADUs as a binary task: classifying sentences as argument or non-argument.
UKPS was obtained with crowd-sourced annotations. 
WEBIS was automatically obtained from debate forums. 
ARAUC was annotated following a comprehensive argumentation theoretical background \cite{walton2009argumentation}.
We discarded the topic that accompanied the UKPS arguments and retrieved the ARAUC corpus as in \cite{Rooney2012ApplyingKM}.
ARAUC and WEBIS were under-sampled to balance the classes.\footnote{The appendices contain details regarding the task, data sets and classifiers necessary for reproduction.}

\textbf{Pre-trained semantic spaces} - Different lexical semantic families are conceived with different semantic relations and objectives, using them as pre-trained LMs may capture idiosyncratic linguistic knowledge useful for the identification of ADUs.
We considered three major lexical semantic families: semantic networks, feature-based models and semantics spaces.

Semantic networks, as the WordNet \cite{fellbaum1998wordnet}, represent lexical semantics with graphs where the nodes are the lexical units and the labeled edges represent semantic relations.
Feature-based models, as the Small World of Words (SWOW) \cite{de2019small}, represent lexical semantics with hash tables where the keys are the lexical units and the keys values represent the semantic relations.
Semantic spaces, as the Word2vec \cite{mikolov2013distributed}, represent lexical semantics with vectors of real value and the semantic relations are extracted using the cosine distance between lexical units.
Different semantic theories guide the creation of these lexical semantic families.
WordNet was obtained with the elaborated work of linguists that reflect on the semantic properties of words.
SWOW was obtained by asking common native speakers of a language to refer to three words they associate when exposed to a cue word.
Word2vec and semantic spaces in general are obtained from corpora, extracting semantic knowledge using the co-occurrence of words.

Both semantic networks and feature-based models encode commonsense knowledge in their relations. 
This makes them candidates for obtaining improvements in a task such as ADUs that requires commonsense knowledge.
Semantic spaces are the typical go-to lexical models given the state-of-art results found across many natural language processing tasks.

As semantic spaces, we used the original Word2vec as also enriched semantic spaces, namely: GloVe, FastText, Dependency, BERT and T5.
GloVe \cite{pennington2014glove} is a count-based method that trains on aggregated global word-word co-occurrence statistics from a corpus. 
The FastText \cite{joulin2016fasttext} model uses morphological information.
The Dependency \cite{levy2014dependency} uses the dependency parse context to guide the co-occurrence train.
BERT \cite{devlin2019bert} processes words in relation to all the other words in a sentence.
T5 \cite{raffel2019exploring} uses a unified text-to-text-format for the input/output resorting to an encoder-decoder architecture.


To compare the different semantic families, we resorted to the conversion of a semantic network (WordNet) and a feature-based model (SWOW) to separate semantic spaces.

The semantic space for WordNet and SWOW followed the conversion of different lexical semantic families to semantic spaces provided in \cite{branco-etal-2020-comparative}. 
We used the graph-based WordNet-RandomWalk and the feature-based SWOW-RandomWalk conversion, given the consistency and performance obtained in their extrinsic evaluation compared with other conversion methods.

\textbf{Classifiers} - We used a BiLSTM~\cite{hochreiter1997long} neural network, a simple yet effective model widely used in previous studies.
The semantic spaces were transferred to the neural network input layer, mapping lexical units to semantic vectors. 
For the contextualized representations we used a fine-tuned BERT \cite{devlin2019bert} and T5 \cite{raffel2019exploring}.\footnote{Using a different classifier, a Transformer architecture \cite{vaswani2017attention}, for the contextualized representations weakens the comparability across pre-trained models but the alternative technique of feature extraction from BERT or T5 would undermine the model by tilting it to a specific layer and thus capturing a narrow range of linguistic phenomena.}
For baselines we used a linear model, support-vector machine (SVM)~\cite{cortes1995support}, and a randomly initialized BiLSTM neural network.

\section{Results and Discussion}

\begin{table*}[]
\centering
\begin{tabular}{llllllllllllllll}
\hline
\multicolumn{1}{c}{\multirow{2}{*}{\textbf{Model}}} & \multicolumn{5}{c}{\textbf{UKPS}} & \multicolumn{5}{c}{\textbf{WEBIS}} & \multicolumn{5}{c}{\textbf{ARAUC}} \\ \cline{2-16} 

\multicolumn{1}{c}{} &
  \multicolumn{1}{c}{\textbf{{\scriptsize P}}} &
  \multicolumn{1}{c}{\textbf{{\scriptsize R}}} &
  \multicolumn{1}{c}{\textbf{{\scriptsize F1}}} &
  \multicolumn{1}{c}{\textbf{{\scriptsize F*}}} &
  \multicolumn{1}{c}{\textbf{{\scriptsize OOV}}} &
   
  \multicolumn{1}{c}{\textbf{{\scriptsize P}}} &
  \multicolumn{1}{c}{\textbf{{\scriptsize R}}} &
  \multicolumn{1}{c}{\textbf{{\scriptsize F1}}} &
  \multicolumn{1}{c}{\textbf{{\scriptsize F*}}} &
  \multicolumn{1}{c}{\textbf{{\scriptsize OOV}}} &
   
  \multicolumn{1}{c}{\textbf{{\scriptsize P}}} &
  \multicolumn{1}{c}{\textbf{{\scriptsize R}}} &
  \multicolumn{1}{c}{\textbf{{\scriptsize F1}}} &
  \multicolumn{1}{c}{\textbf{{\scriptsize F*}}} &
  \multicolumn{1}{c}{\textbf{{\scriptsize OOV}}} \\ \hline
SVM                                                 & .73 & .73 & .72 &     & .03 & .70 & .69 & .69 &     & .05  & .54 & .54 & .53 &     & .09 \\
Random                                              & .67 & .78 & .72 & .72 & .04 & .65 & .70 & .68 & .70 & .06  & .52 & .64 & .56 & .55 & .11 \\ \hline
WordNet                                             & .67 & .79 & .71 & .69 & .22 & .54 & .71 & .61 & .64 & .22  & .51 & .62 & .55 & .56 & .26 \\
SWOW                                                & .71 & .76 & .73 & .72 & .08 & .60 & .84 & .70 & .68 & .09  & .50 & .63 & .55 & .54 & .16 \\ \hline
Word2vec                                            & .73 & .78 & .75 & .72 & .19 & .63 & .78 & .70 & .68 & .19  & .49 & .65 & .56 & .54 & .23 \\
GloVe                                               & .73 & .74 & .74 & .72 & .03 & .69 & .77 & .73 & .73 & .06  & .52 & .67 & .58 & .57 & .11 \\
Fasttext                                            & .73 & .78 & .75 & .72 & .04 & .71 & .75 & .73 & .71 & .06  & .51 & .62 & .56 & .55 & .11 \\
Dep.                                          		& .71 & .77 & .74 & .72 & .06 & .69 & .80 & .74 & .72 & .06  & .51 & .64 & .57 & .54 & .12 \\ \hline
BERT                                                & .87 & .70 & \textbf{.77} &     &     & .89 & .95 & .92 &     &      & .70 & .70 & .68 &     &     \\
T5                                                  & .74 & .80 & \textbf{.77} &     &     & .98 & .98 & \textbf{.98} &     &      & .81 & .67 & \textbf{.74} &     &     \\ \hline

\end{tabular}
\caption{Results for all the experiments. \textbf{P}=Precision; \textbf{R}=Recall; \textbf{F1}=F1 measure; \textbf{F*}=F1 measure with a trainable semantic space (weights are not frozen). \textbf{OOV}=Out-of-vocabulary words.}
\label{table:results}
\end{table*}

The experiments results can be found in Table \ref{table:results}.
Several models surpassed the baselines (F1).
BERT and T5 surpassed all models.
Although obtaining top performance was not the aim of this work, the results are in line with state-of-the-art results found in \cite{schulz2018multi} for UKPS, \cite{Khatib2016CrossDomainMO} for WEBIS and \cite{moens2007automatic,palau2009argumentation,Rooney2012ApplyingKM} for ARAUC. 

When comparing the different semantic families there is no dominance of one of these models over the others, with one notable exception, contextual semantic spaces.
It is worth mentioning that WordNet and SWOW have a smaller training data set and subsequently a higher OOV. 
Nevertheless, they still obtain results close to the other models, suggesting that larger data sets may increase the scores.
The scores also indicate that converting semantic networks (WordNet) and feature-based models (SWOW) fine-tuned to semantic similarity and relatedness tasks works for the identifications of ADUs.
Although enriched semantic spaces encode a higher level of linguistic complexity these only improved slightly over the vanilla Word2vec model.
The examples available for ARAUC were insufficient to obtain convergence with the BiLSTM models. 
The results also show that, in general, training with a fixed semantic space versus a trainable semantic space (F1 vs F*) improves the scores corroborating that transfer learning the knowledge of semantic spaces leverage the score.

\section{Analysis}
To analyze: a) the impact of the data size, we report on a profile study; b) the diverseness of the data sets constituent structure, we perform a cross-train; c) the interpretability of the lexical features, we analyze by proxy the bag-of-words used as features in the SVM; d) the complementary of the models, we evaluate a range of ensembles; and finally, e) the task linguistic complexity, we evaluate the models with downstream tasks encompassing different fine distinctions of lexical and linguistic organization.

\textbf{Profiling analysis} - Evaluating the models with incremental training sizes shows that in general, all models improve with more data (F1). 
The exception goes for all non-contextual models trained on ARAUC. When using all the ARAUC data the scores are two points below the top-performing score.
When comparing the train with half versus full UKPS data the scores only improved two points.
This points out to a possible plateau and the need for more extrinsic knowledge.

\textbf{Cross-training} - We experimented training iteratively with a data set and testing on the left-out data sets.
All models obtained a random accuracy, except for the models trained with an SVM/BERT on UKPS and evaluated on WEBIS. 
Although this suggests that UKPS data can be generalized to WEBIS, the results show a high degree of heterogeneity across data sets.

\textbf{BoW analysis} - To perform a bag-of-words analysis we intersected the SVM features obtained across the data sets (unigrams and bigrams).
As expected, a quantitative analysis of the content versus functional unigrams shows that content words are five times more frequent relevant features.
We expected to find signal words but no obvious patterns or data bias were identified across the data sets.
This seems to indicate that grammatical structure and context play a major role in solving the identification of ADUs.

\textbf{Ensemble} - We experimented with several model ensembles using the largest data set (UKPS).
All the ensembles obtained worst results except a weighted ensemble with marginal improvements.\footnote{Ensemble obtained with BERT, SVM and the Dependency model improving +0.003 F1.}
We hypothesized that knowledge from different semantic families as WordNet or SWOW would complement other semantic spaces but this was not shown on the ensemble scores.

\begin{figure*}[h]
  \centering
  \includegraphics[scale=0.8]{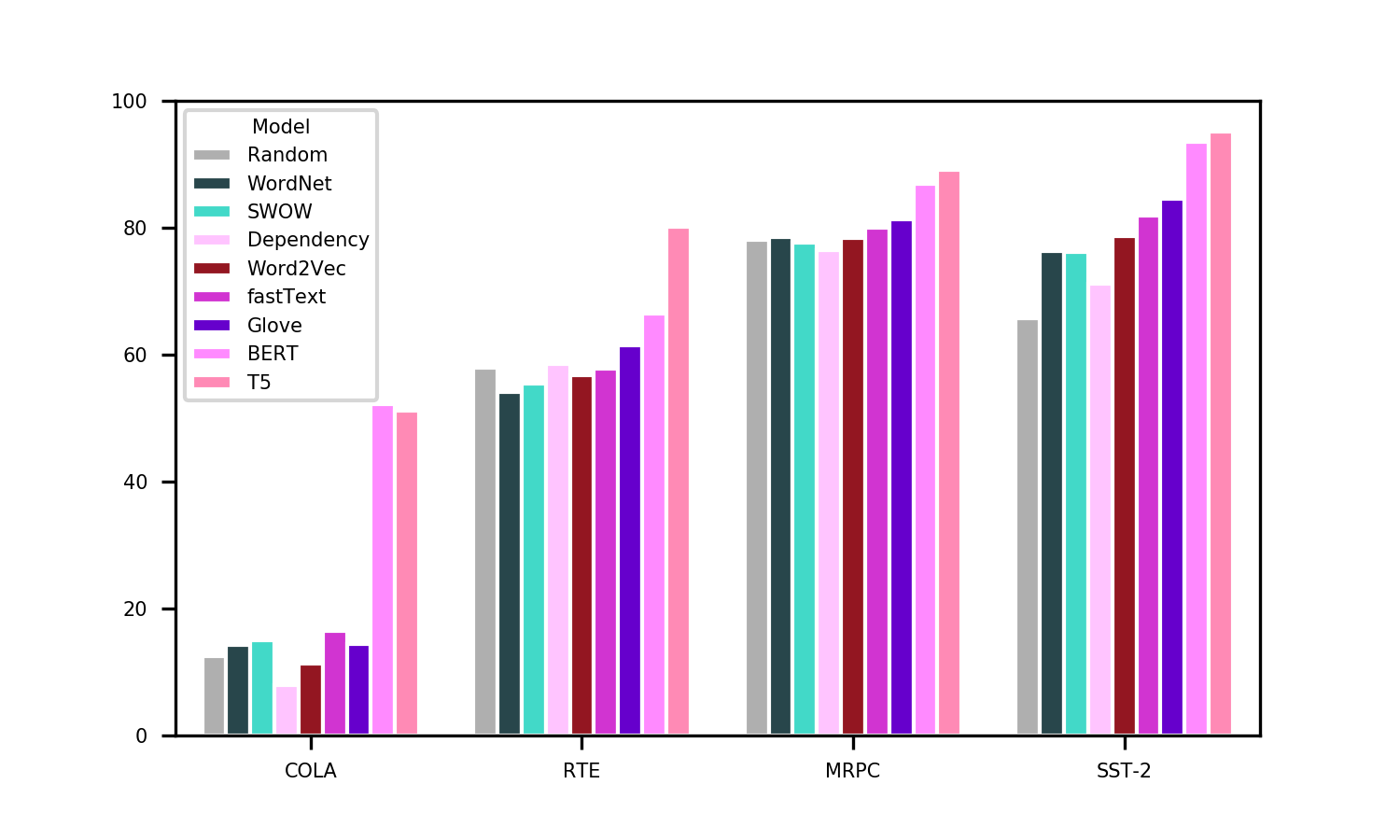}
  \vspace{-0.5cm}
  \caption{\label{figure:extrinsic_intermediate}\textbf{Linguistic complexity evaluation}: Performance of models with different semantic spaces (bars) over four different GLUE \textbf{downstream tasks} (groups), measured in different evaluation metrics projected to a common range [0, 100]. The reported scores are the average of three runs.}
\end{figure*}

\textbf{Parallel with fine-grained downstream tasks} - 
To assert a possible range of linguistic complexity addressed by the semantic spaces we resort to the evaluation of parallel downstream tasks.
These tasks benefit in different degrees of fine-grained information regarding the lexical knowledge required to address them, some with a fine distinction between the meaning of the different words.
We resorted to four tasks of GLUE \cite{wang2018glue}, a standard benchmark for natural language understanding tasks. 
For the classifier, we adopted the BiLSTM encoder from \cite{wang2019can} for comparability with existing semantic spaces evaluated on the tasks, except with the contextualized models where we used the fine-tuned version for each data set.
The following  downstream tasks were used: \textbf{CoLA}, a language membership task (grammatical acceptability); \textbf{RTE}, an entailment task with the data from RTE1-RTE5 (natural language inference); \textbf{MRPC}, a paraphrase task (semantic equivalence) and \textbf{SST-2}, a sentiment analysis task. 
The results are plotted in Figure \ref{figure:extrinsic_intermediate}.
Arranging the performance of the tasks from worst to best, as in Figure \ref{figure:extrinsic_intermediate}, provides several insights: higher linguistic complexity downstream tasks are harder to solve, lexical semantic spaces achieve high performance on low linguistic complexity but fail to address harder tasks and contextualized semantic spaces are successful across tasks.  

In the CoLA task, some models perform below or at best on a par with a random semantic space.
The task is very hard as it relies on rich information about the grammatical structure of the sentences and to categorically decide sentence membership in the language defined by the grammar.
Hence, the evaluation scores seem to indicate that the signal from the lexical information encoded in pre-trained models, from whatever lexical theory or empirical source, has a somewhat marginal impact. 
While the fine-tuned contextual models can address CoLA, the same parallel can be observed with the ADUs task, this correlation suggests that highly complex linguistic information plays a major role to solve the task.

\section{Conclusion}

This paper addresses a clear point, to measure the impact of models pre-trained with different lexical semantic families on ADUs identification.
We presented a focused contribution resorting to three different and heterogeneous data sets, with the transfer learning of semantic spaces from different semantic families.
Using transfer learning techniques proved to be beneficial to the task.
When proposing that in some way or the other the correlation of downstream tasks mirrors the linguistic complexity found in the ADUs task, the results obtained suggest that ARAUC encodes higher linguistic complexity followed by UKPS and finally by an almost completely lexical dependent WEBIS data set. 
We expected the contribution of the commonsense knowledge encoded in different lexical semantic models.
The somewhat negative results using different lexical semantic families and the lack of improvements using ensembles points the future work to alternatives for the training-objectives of current LMs architectures, for example exploring neuro-symbolic approaches guided with feature-based models and semantic networks. 

\newpage
\bibliography{bibliography}
\bibliographystyle{acl}

\newpage
\begin{appendices}

The following appendices include information that allow the reproduction of the paper along with non-essential detailed information about the data sets (Appendix \ref{appendix_datasets}), conversion process (Appendix \ref{appendix_conversion}), semantic spaces (Appendix \ref{appendix_semanticspaces}), classifiers (Appendix \ref{appendix_classifiers}) and experiments results (Appendix \ref{appendix_results}).

\section{Data sets}
\label{appendix_datasets}
Next, it follows a detailed description of each data set used in the evaluation process.
An example of each data set is shown in Table \ref{table:datasets_examples}.
In Table \ref{table:datasets} the number of sentences and tokens are quantified.
We split the data sets in 70\% for the train, 10\% for the development and 20\% for the testing. 

\textbf{- UKP Sentential Argument Mining Corpus} (UKPS) \cite{Stab2018CrosstopicAM}.
UKPS was created from eight controversial topics and across domains: news reports, editorials, blogs, debate forums, and encyclopedia articles.
Each sentence in the corpus was manually annotated (concerning the topic) as argument or non-argument.
The corpus has approximately 25k sentences, 10k arguments and 15k non-arguments.
An argument is defined as \textit{a span of text that conveys evidence or reasoning that can be used to either support or oppose a given topic}.

\textbf{- Webis Debate 16} (WEBIS) \cite{Khatib2016CrossDomainMO}.
WEBIS was created from an online-debate platform and obtained with a distant supervision algorithm.
The meta-data from the debates was mapped to units of argument or non-argument discourse.
Meta-data referring to the debate \textit{introduction} was labeled as non-argument, the meta-data for \textit{points for or against} and \textit{counterpoints} were labeled as arguments.
The corpus has approximately 30k sentences, 24k arguments and 5k non-arguments. 
We under-sampled the data set to balanced it, resulting in a total of 11,112 examples.

\textbf{- AraucariaDB} (ARAUC) \cite{Reed2008LanguageRF}.
ARAUC was created from different domains: news editorials, parliamentary records, judicial summaries and discussion boards. 
The corpus uses the Walton \cite{walton2009argumentation} argumentative scheme set theory and is originally stored in a graph structure.
As in \cite{Rooney2012ApplyingKM}, all the sentences not containing part of a \textit{premise} or \textit{conclusion} were considered a non-argument.
The corpus was under-sampled to balanced it, it resulted in 3k sentences.

\begin{table*}[!htbp]
\begin{tabular}{p{1.4cm}p{11cm}p{2.2cm}}
\hline
\textbf{Data set}                        & \textbf{Example}                                                                         & \textbf{Label} \\ \hline
\multicolumn{1}{l}{\multirow{2}{*}{UKP}} & We need a safe, genuinely sustainable, global and green solution to our energy needs, not a dangerous diversion like nuclear power.                 & argument       \\
\multicolumn{1}{c}{}                     & There are many notable authors of books and articles that render scientific findings available in lay language to a wider public.    & non-argument   \\ \hline
\multirow{2}{*}{WEBIS}                   & Having the whole of an ethnicity within one state will help prevent misappropriation of culture and history by another state. & argument       \\
                                         & Since the end of the AU’s peace and security council has had responsibility for maintaining peace in Africa.    & non-argument   \\ \hline
\multirow{2}{*}{ARAUC}                   & To build a more hopeful America we must help our children reach as far as their vision and character can take them.                      & argument       \\
                                         & To emphasize the priority of youth affairs the king answered questions from three college students via an international webcast.                 & non-argument   \\ \hline
\end{tabular}
\caption{Examples of what is labeled as argument or non-argument for each of the data sets.}
\label{table:datasets_examples}
\end{table*}

\begin{table}[!htbp]
\centering
\begin{tabular}{lrr}
\hline
\textbf{Data set} & \multicolumn{1}{l}{\textbf{Sentences}} & \multicolumn{1}{l}{\textbf{Tokens}} \\ \hline
UKPS              & 25,492                                   & 609,740 (23.92)     \\
WEBIS             & 11,112                                   & 297,635 (26.51)     \\
ARAUC             & 3,000                                    & 53,091 (17.70)      \\ \hline
\end{tabular}
\caption{Argument identification data sets (after undersampling). The \textbf{Sentences} column show the total of arguments/non-argument. The \textbf{Tokens} column show the total of tokens and tokens per sentence.}
\label{table:datasets}
\end{table}

\section{Lexical semantic families conversion to semantic spaces}
\label{appendix_conversion}

The typical representation factors for each of the lexical semantic families can be seen in Table \ref{table:semantic_families}.
For the conversion from the WordNet, a Semantic Network, to a semantic space the nodes were projected to a raw corpus. 
Each node was written to the corpus (lemma) followed by all the nodes with corresponding relations.
For the conversion from SWOW, a Feature-based model, to a semantic space the keys lexical represention (word) were projected to a raw corpus. 
Each key was written to the corpus followed by all the key's association cues.

We used the Skip-gram algorithm to create a semantic space for each corpora.
We resorted to the Gensim Skip-gram implementation \cite{rehurek_lrec}.
We fixed the vector dimension at 300 units and performed a grid-search on four hyper-parameters: window (3, 10, 20), negative samples (5, 25, 100), learning rate (0.1, 0.01, 0.001) and epochs (1, 5, 10).
The final semantic spaces for WordNet and SWOW obtained 0.5389 and 0.5261 Spearman rank-order correlation coefficient in Simlex, and with WordSim352-Rel obtained 0.6719 and 0.4541 Spearman rank-order correlation coefficient.
For comparison, the word2vec obtains 0.4361 and 0.6989  Spearman rank-order correlation coefficient in Simlex and WordSim352-Rel respectively.

\begin{table}[!htbp]
\centering
\begin{tabular}{lllll}
\hline
\textbf{Semantic Family} & \textbf{Representation} & \textbf{Lexical Unit} & \textbf{Semantic Relations} & \textbf{Example}     \\ \hline
Semantic Networks        & graph                   & node                  & edge                        & WordNet              \\
Feature-based models     & hash tables             & key                   & key values                  & SWOW \\
Semantic spaces          & vector space            & vector                & cosine distance             & word2vec             \\ \hline
\end{tabular}
\caption{Lexical semantic families and their typical representation factors. The Semantic relations column refers to how the semantic similarity of two different words is measured.}
\label{table:semantic_families}
\end{table}

\section{Semantic spaces}
\label{appendix_semanticspaces}

Table \ref{table:dsms} summarizes the semantic spaces (except contextual semantic spaces).
We used the original Word2Vec model \cite{mikolov2013distributed} and dependency model \cite{levy2014dependency}.
For GloVe we used the 6 Billion token model trained with the Wikipedia 2014 and the Gigaword 5 corpora.
For Fasttext we used the 1 million word vectors trained on Wikipedia 2017, UMBC web base corpus and statmt.org news data set.

\begin{table}[]
\centering
\begin{tabular}{llrr}
\hline
\textbf{Model} & \textbf{Domain}   & \multicolumn{1}{l}{\textbf{\#Vocab.}} & \multicolumn{1}{l}{\textbf{\#Dim}} \\ \hline
Word2Vec       & News              & 3,000,000                             & 300                                \\
GloVe          & Wiki. \& News     & 400,000                               & 300                                \\
Fasttext       & Wiki. \& News     & 999,994                               & 300                                \\
Dependency           & Wiki.             & 174,015                               & 300                                \\ \hline

\end{tabular}
\caption{Summary of semantic spaces models regarding the domain of the data used for training, the resulting vocabulary of the model and the word vector dimension.}
\label{table:dsms}
\end{table}

For the contextual semantic space we used the original BERT models fine-tuned to the task data sets.
We fine-tuned the BERT-Base Uncased (uncased\_L-12\_H-768\_A-12), on each of the three data sets (UKPS, WEBIS and ARAUC).
We performed a grid-search on 2 hyper-parameters: epoch (10, 8, 5, 3, 1) and learning rate (0.002, 0.0002, 0.00002, 0.000002, 0.000002).
We used a maximum sequence length of 128 units and a 32 train batch size for all of the models. 
Given that BERT and T5 use a word segmentation method we can assume that there are no out-of-vocabulary words.

\section{Classifiers}
\label{appendix_classifiers}

The hyper-parameters were obtained from a grid-search using a random (vectors) semantic space baseline (a strong baseline given that it was the only model, along with contextual models where the hyper-parameters were tuned). 
For the vocabulary of the baseline we used the 1M vocabulary extracted from Fasttext (only lexical units, random vectors were generated).
For the UKPS and WEBIS development set the top performing hyper-parameters were:
10 epochs, a batch size of 64, a sequence length of 30, a learning rate of 0.01, a dropout of 0.8 and 48 LSTM units.
For the ARAUC development set the top performing hyper-parameters were:
20 epochs, batch size of 16, a sequence length of 15, a learning rate of 0.001, a dropout of 0.7 and 1 LSTM unit.
All the models were trained 10 times and the scores presented are the average of all the runs.

\section{Results}
\label{appendix_results}

The values for the extrinsic evaluation and used for plotting Figure \ref{figure:extrinsic_intermediate} are more detailed in Table \ref{table:extrinsic_intermediate}.

\begin{table*}[htb!]
    \centering
    \begin{tabular}{lllll}
    \cline{2-5}
    \multicolumn{1}{l}{} &
    \multicolumn{1}{c}{CoLA} & \multicolumn{1}{c}{RTE} & \multicolumn{1}{c}{MRPC} & \multicolumn{1}{c}{SST-2}  \\
    \hline
    Random      & 12.37 $\pm$ 0.65 & 57.90 $\pm$ 2.35 & 78.02 $\pm$ 0.78 & 65.70 $\pm$ 0.95     \\ \hline\hline
    WordNet & 14.13 $\pm$ 0.26 & 54.07 $\pm$ 0.74 & 78.47 $\pm$ 0.77 & 76.37 $\pm$ 0.77 \\
    SWOW & 14.97 $\pm$  1.17 & 55.37 $\pm$ 2.39 & 77.60 $\pm$ 1.07 & 76.20 $\pm$ 0.37 \\ \hline\hline
    Dependency  & 07.80 $\pm$ 2.41  & 58.47 $\pm$ 2.32  & 76.50 $\pm$ 0.43 & 71.17 $\pm$ 1.22     \\
    Word2Vec & 11.33 $\pm$ 0.68 & 56.70 $\pm$ 0.0 & 78.40 $\pm$ 1.21 & 78.73 $\pm$ 0.25  \\
    fastText & 16.43 $\pm$ 2.25 & 57.73 $\pm$ 1.56 & 79.92 $\pm$ 0.48 & 81.87 $\pm$ 0.76 \\
    Glove       & 14.40 $\pm$ 0.78  & 61.47 $\pm$ 2.56  & 81.30 $\pm$ 0.23 & 84.57 $\pm$ 0.58     \\
    BERT & \textbf{52.10} & 66.40 & 86.85 & 93.50 \\
    T5 & 51.10 & \textbf{80.10} & \textbf{89.10} & \textbf{95.20} \\
    \hline
    \end{tabular}
    \caption{\label{table:extrinsic_intermediate}\textbf{Extrinsic evaluation}: Performance over five GLUE \textbf{downstream tasks} (columns) of models with different input layer word embeddings (rows). For the task CoLA, performance is measured with Matthews correlation. For MRPC, an average of accuracy and F1 is reported. For the remaining tasks, accuracy is reported. The evaluation scores were projected to a [0-100] common scale (higher is better), with bold denoting top results. Each score is the average of the results from three runs with the random seeds 1147, 1256 and 1179. To enhance the readability of eventual data patterns, the content of this table is rendered in Figure \ref{figure:extrinsic_intermediate}.}
\end{table*}

\end{appendices}

\end{document}